\pdfoutput=1
\PassOptionsToPackage{table,dvipsnames,svgnames}{xcolor}
\documentclass[11pt]{article}

\usepackage[final]{acl}
\usepackage{natbib}
\usepackage{algorithm}
\usepackage{algpseudocode}
\usepackage{amsmath}
\usepackage{times}
\usepackage{latexsym}
\usepackage{amssymb}
\usepackage{xcolor}
\usepackage[T1]{fontenc}

\usepackage[utf8]{inputenc}

\usepackage{microtype}
\usepackage{makecell}
\usepackage{inconsolata}
\usepackage{multirow}     
\usepackage{graphicx}     
\usepackage{makecell}     
\usepackage{booktabs}     
\usepackage{paralist}
\usepackage{footmisc}

\usepackage{graphicx}
\DeclareMathOperator*{\argmax}{arg\,max}

\title{Reviving Cultural Heritage: A Novel Approach for Comprehensive Historical Document Restoration}

\author{
 \textbf{Yuyi Zhang\textsuperscript{1,3}}~~
 \textbf{Peirong Zhang\textsuperscript{\footnotemark[2] 1}}~~
 \textbf{Zhenhua Yang\textsuperscript{\footnotemark[2] 1}}~~ 
 \textbf{Pengyu Yan\textsuperscript{1,3}}~~  
 \textbf{Yongxin Shi\textsuperscript{1}}~~  
\\
 \textbf{Pengwei Liu\textsuperscript{2,3}}~~  
 \textbf{Fengjun Guo\textsuperscript{2,3}}~~  
 \textbf{Lianwen Jin\textsuperscript{\footnotemark[1]~1,3,4}}
\\
\\
 \textsuperscript{1}South China University of Technology
\\ \textsuperscript{2}Intsig Information Co., Ltd.
\\ \textsuperscript{3}INTSIG-SCUT Joint Lab on Document Analysis and Recognition
\\ \textsuperscript{4}SCUT-Zhuhai Institute of Modern Industrial Innovation
\\
 \small{
   \href{yuyi.zhang11@foxmail.com}{yuyi.zhang11@foxmail.com}~~
   \href{eelwjin@scut.edu.cn}{eelwjin@scut.edu.cn}
 }
}

\begin{document}
\maketitle
\renewcommand{\thefootnote}{\fnsymbol{footnote}}
\footnotetext[2]{Equal contribution}
\footnotetext[1]{Corresponding authors.}
\renewcommand{\thefootnote}{\arabic{footnote}}

\begin{abstract}

Historical documents represent an invaluable cultural heritage, yet have undergone significant degradation over time through tears, water erosion, and oxidation. 
Existing Historical Document Restoration (HDR) methods primarily focus on single modality or limited-size restoration, failing to meet practical needs. To fill this gap, we present a full-page HDR dataset (\textbf{FPHDR}) and a novel automated HDR solution (\textbf{AutoHDR}). Specifically, FPHDR comprises 1,633 real and 6,543 synthetic images with character-level and line-level locations, as well as character annotations in different damage grades. 
AutoHDR mimics historians' restoration workflows through a three-stage approach: OCR-assisted damage localization, vision-language context text prediction, and patch autoregressive appearance restoration.
The modular architecture of AutoHDR enables seamless human-machine collaboration, allowing for flexible intervention and optimization at each restoration stage.
Experiments demonstrate AutoHDR's remarkable performance in HDR. 
When processing severely damaged documents, our method improves OCR accuracy from 46.83\% to 84.05\%, with further enhancement to 94.25\% through human-machine collaboration.
We believe this work represents a significant advancement in automated historical document restoration and contributes substantially to cultural heritage preservation.
The model and dataset are available at \href{https://github.com/SCUT-DLVCLab/AutoHDR}{https://github.com/SCUT-DLVCLab/AutoHDR}.
\end{abstract}

\section{Introduction}

Historical documents, encompassing books, rubbings, scrolls, and inscriptions, stand as a vital window into ancient civilizations and wisdom. Through the ages, they have sustained deterioration from various environmental factors, such as improper storage, transportation, and wartime upheavals, resulting in physical damage, water erosion, and oxidation. Therefore, restoring these ancient treasures is crucial to preserving their cultural and historical significance.

\begin{figure}[t]
  \includegraphics[width=1\columnwidth]{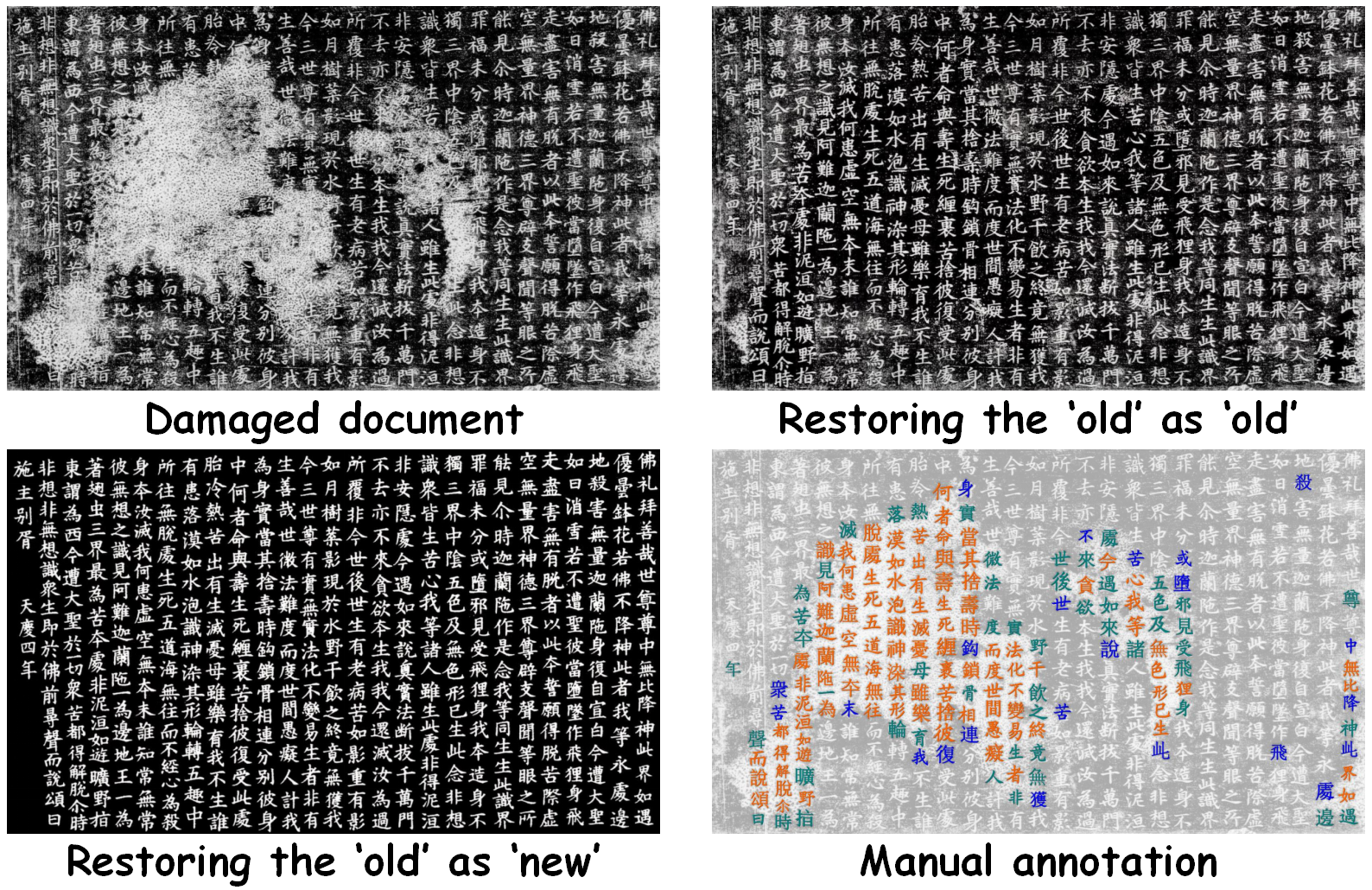}
  \centering
  \vspace{-8mm}
  \caption{Restoration results of our AutoHDR. 
  \textcolor{orange}{Orange}, \textcolor{green}{green}, and \textcolor{blue}{blue} indicate severe, medium, and light damage, respectively.
  }
  \label{fig: rs}
  \vspace{-7mm}
\end{figure}

\begin{figure*}[t]
  \includegraphics[width=0.98\linewidth]{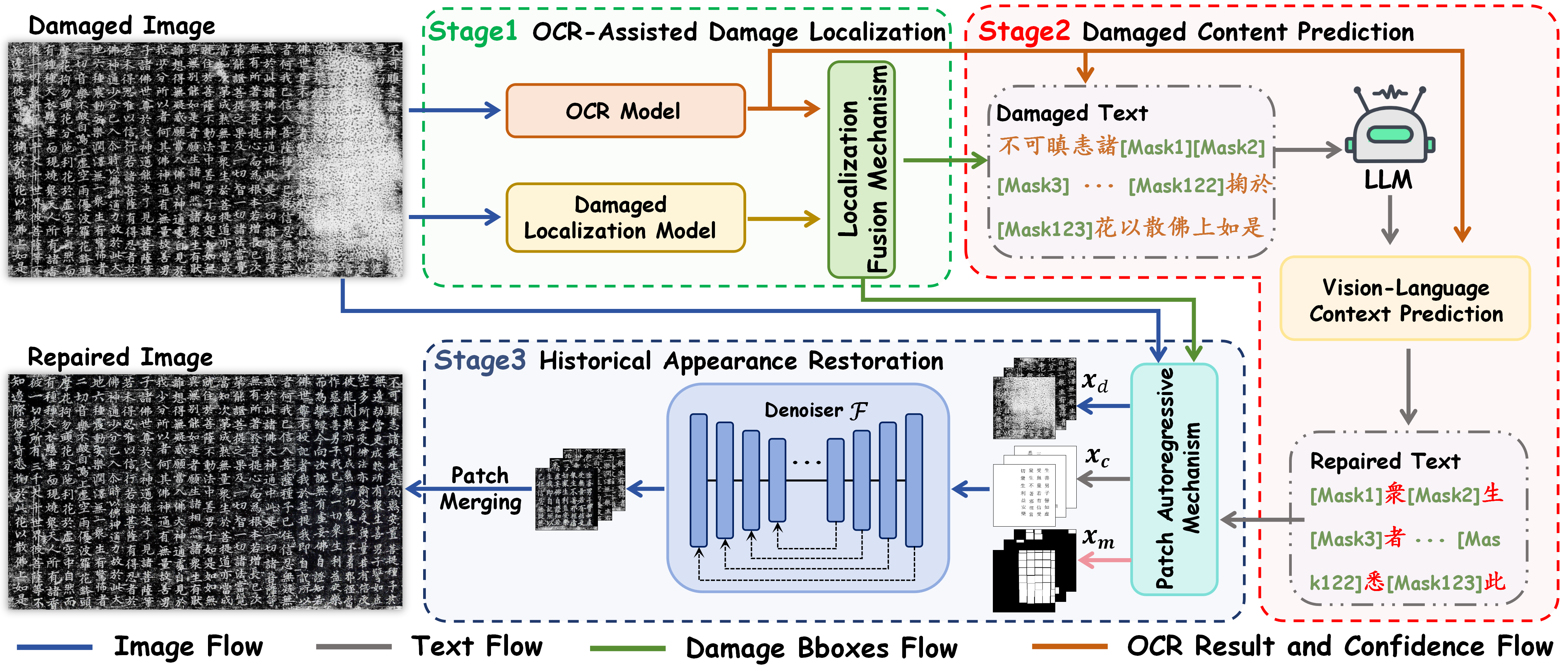}
  \centering
  \vspace{-3mm}
  \caption{Overall workflow of the proposed AutoHDR. The framework contains three distinct yet interconnected stages: OCR-Assisted Damage Localization for character recognition and damage localization, Damaged Content Prediction for text restoration, and Historical Appearance Restoration for pixel-level reconstruction.}
  \label{fig: pipe}
  \vspace{-5mm}
\end{figure*}

Yet, the task of Historical Document Restoration (HDR), remains complex and time-consuming. Traditional manual restoration involves three key stages: (1) identifying damaged regions through specialized knowledge and historical literature, (2) reconstructing damaged content based on literature references, and (3) applying delicate conservation techniques to restore the documents' original appearance. To alleviate the huge labor cost, various automated HDR techniques have been proposed. For example, \citet{Ithaca} employs a Transformer to predict damaged text, geographic origins, and dates. \citet{diffhdr} restores historical appearances through manually provided annotations. However, existing methods confront several critical limitations. (1) Most methods are confined to single-modal restoration (text- or image-only). (2) While some multimodal approaches are proposed, they are restricted to processing damage in very small regions, such as single image patches or a few characters. (3) The limited perceived region leads to two cascading problems: i) models fail to leverage broader contextual semantic information, and ii) restoration often fails when the damage extends beyond the model's processing patch size. (4) Most importantly, current methods only address single stages of the restoration process, failing to provide a fully automated solution for all stages. Thus, manual intervention is still required in the restoration procedure, preventing the complete liberation of human experts from this demanding workload.

To address these challenges, we propose \textbf{AutoHDR}, a novel fully \underline{Auto}mated solution for full-page \underline{HDR}. As shown in Fig.~\ref{fig: pipe}, AutoHDR mirrors the workflow of historians, jointly restoring text and historical appearance in three stages: OCR-assisted damage localization, damaged content prediction, and historical appearance restoration. By combining OCR-assisted visual information with the language understanding capabilities of LLMs, AutoHDR achieves precise localization and restoration of damaged text. Then, adhering to the principle of "restoring the old as the old\footnote{This means that the repaired text and background appear consistent with the condition of the ancient document.}" \cite{BJJG199904026, TSGL202107023}, we design a patch autoregressive restoration approach to reconstruct the original document appearance at the page level, which conducts progressive restoration from simple to complex cases to ensure high fidelity. To our surprise, as depicted in Fig.~\ref{fig: rs}, AutoHDR not only excels at "restoring the old as old", but also extends to "restoring the old as new\footnote{This means that the repaired text and background appear as if they are in a new and pristine condition.}", providing higher flexibility to users. Note that the whole procedure is completely automatic, eliminating the need for human intervention. 
Furthermore, the modular architecture of AutoHDR enables seamless human-machine collaboration, allowing for flexible intervention and optimization at each restoration stage.

Subsequently, given the scarcity of HDR datasets and the limited focus on patch-level restoration of existing ones \cite{zhumm, diffhdr}, we introduce \textbf{FPHDR}, a pioneer dataset for \underline{F}ull-\underline{P}age \underline{HDR}. It includes 1,633 expertly annotated real samples and 6,543 high-quality synthetic samples, each providing character- and line-level location as well as character annotations in different damage grades, serving as a comprehensive benchmark for HDR model training and evaluation.

Extensive experiments are conducted to evaluate AutoHDR's performance, which reveals its remarkable advantages over existing methods in both text restoration accuracy and historical appearance preservation. For severely damaged documents where OCR recognition accuracy starts at merely 46.83\%, AutoHDR substantially improves the accuracy to 84.05\%. Moreover, when combined with expert collaboration, the accuracy further rises to 94.25\%. These compelling results not only validate AutoHDR's effectiveness as a standalone system but also underscore its potential as a powerful assistive tool for historians in practical applications.

We outline our main contributions as follows:
\begin{itemize}
    \item We propose a novel fully \underline{Auto}mated solution for \underline{HDR} (AutoHDR), inspired by mirroring the workflow of expert historians.
    \item We introduce a pioneer \underline{F}ull-\underline{P}age \underline{HDR} dataset (FPHDR), which supports comprehensive HDR model training and evaluation. 
    \item Extensive experiments demonstrate the superior performance of our method on both text and appearance restoration.
    \item The modular design enables flexible adjustments, allowing AutoHDR to collaborate effectively with historians.
\end{itemize}

\section{Related Work}
Historical document restoration primarily involves two modalities~\cite{survey}, i.e., text and visual appearance. 

\textbf{Historical Text Restoration}: 
Traditional historical text restoration relies heavily on expert labor, while recent advances in natural language processing (NLP) techniques offer promising solutions for this field.
Pythia~\cite{Pythica} pioneered Greek text restoration at both character and word levels, inspiring text restoration research across various languages~\cite{fetaya2020restoration, bamman2020latin, lazar2021filling, papavassileiou2023generative}. 
Notably, Ithaca~\cite{Ithaca} employs a transformer to jointly predict damaged texts, geographic origins, and dates, leveraging multi-task learning for enhanced performance.

\textbf{Historical Appearance Restoration}: 
Early methods in historical document appearance restoration depended on traditional image processing, such as \citet{hedjam} using ink's properties under spectra to restore documents.
Other methods focused on improving historical documents legibility~\cite{raha2019restoration, cao2022character, wadhwani2021text}. For instance, \citet{cao2022character} introduced adaptive binarization to isolate text from degraded backgrounds.
Deep learning advances have enabled GAN-based~\cite{agtgan, shi2022rcrn} and Diffusion-based~\cite{li2024towards} methods in historical appearance restoration, though mainly for single-character restoration.
Given these limitations, \citet{diffhdr} developed DiffHDR, a patch-level restoration method that preserves the original style but needs manual guidance.

\textbf{Joint Restoration}: 
Recent research has transitioned to restoring historical texts and images jointly.
\citet{lant} introduced a text-appearance restoration method via crowdsourcing, which requires labor input.
\citet{duan} introduced a model that jointly restores degraded texts and images by integrating contextual information but is limited to processing small-scale regions (a few characters).
\citet{zhumm} proposed a restoration framework that performs global appearance restoration followed by text correction through corpus retrieval, then conducts local refinement.
However, this approach relies heavily on corpus coverage and is limited to patch-level binarized images.

\section{FPHDR Dataset}
Current open-source HDR datasets are severely scarce. While datasets like HDR28K~\cite{diffhdr} and CIRI~\cite{zhumm} exist, their restriction to patch-level images prevents the effective utilization of leverage full-page contextual information.
To fill this gap, we introduce FPHDR, a page-level dataset with 1,633 labor-annotated samples for model evaluation and 6,543 synthetic samples for training.

\subsection{Data Collection}
\label{sec: data collection}

The Fangshan Stone Sutras (FSS) is China's largest surviving stone Buddhist canon\footnote{\href{https://en.wikipedia.org/wiki/Chinese_Buddhist_canon}{wikipedia-Chinese Buddhist canon}}. However, extensive damage has hindered research on many sutras, making their restoration both an academic and social imperative.
To address this, we invest substantial effort in collecting 1,633 typical damaged samples from the FSS and manually annotate both their damage locations and damage contents.
However, these data cannot meet the training requirements of HDR models, since diffusion-based appearance restoration models demand pixel-level ground truths, which are impractical to generate manually. 
Therefore, we curate 6,543 well-preserved samples from the FSS, MTHv2~\cite{mth}, and M$^5$HisDoc~\cite{m5} to synthesize pixel-level damaged-restored image pairs as training data.

The collected data exhibit the following characteristics:
\textbf{(1) Semantic Integrity}: All samples maintain complete page-level context, preserving complete contextual semantic information. 
\textbf{(2) Degradation Diversity}: The data features a wide range of typical historical damages, such as surface erosion, radical loss, and character blur, presenting great challenges to HDR models. 
\textbf{(3) Dynasty Diversity}: The collected degradation samples span nearly a millennium from the Sui (581AD-618AD) to the Ming Dynasty (1368AD-1644AD), capturing both character evolution and degradation patterns across history. 
\textbf{(4) Source Diversity}: Various forms of historical documents are considered, including manuscripts, rubbings, and scrolls, representing diverse materials.

\subsection{Manual Annotation for Damage}

\begin{figure}[t]
  \includegraphics[width=0.98\columnwidth]{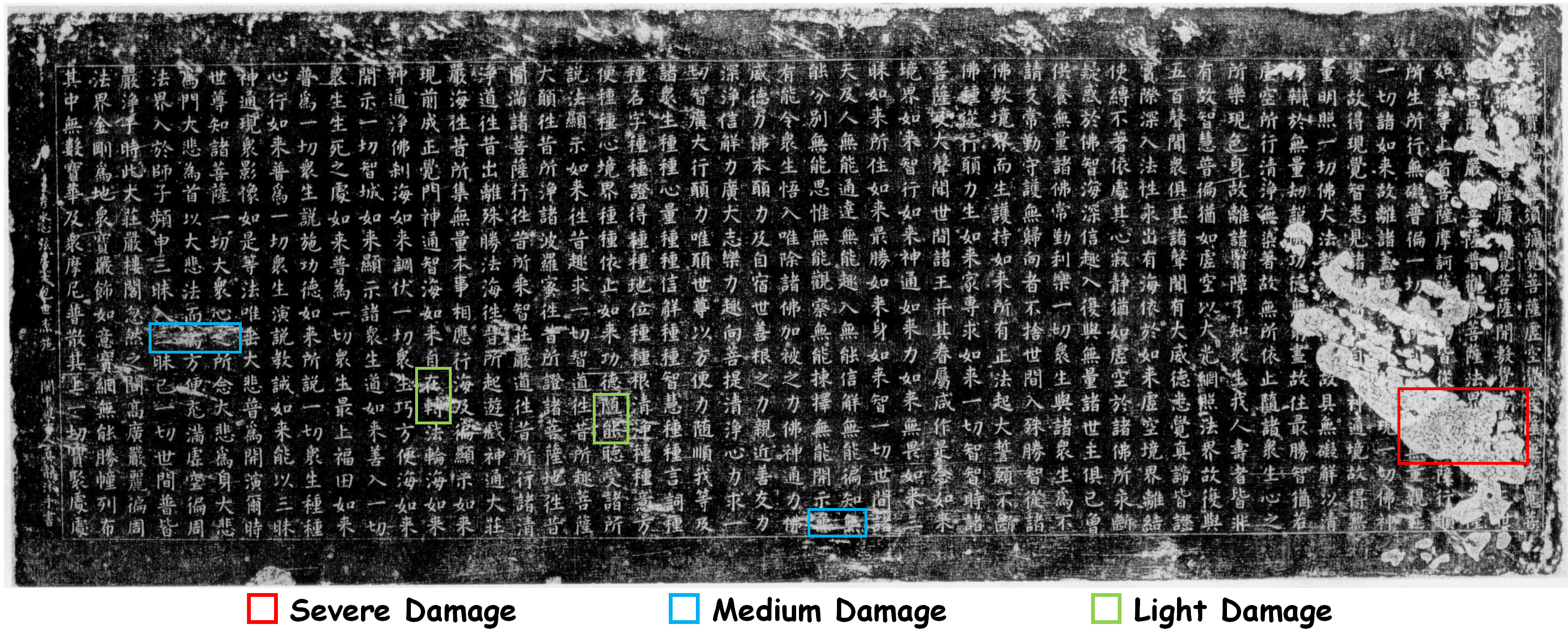}
  \centering

  \caption{Illustration of damage grades in FPHDR. All damages are annotated at the character level, though only typical cases are highlighted here for clarity.}
  \vspace{-2mm}
  \label{fig: grade}
\end{figure}

Due to long-term deterioration, historical documents have sustained varying degrees of damage, rendering their textual content partially or completely illegible. To ensure high-quality annotations of these damaged characters, we curate a professional annotation team consisting of ten experts with over five years of experience in HDR.
Specifically, our annotation process consists of three main steps. 
\textbf{(1) Character Localization}: We annotate bounding boxes for all clearly visible characters and determine the positions of damaged characters based on the layout. 
\textbf{(2) Damage Assessment and Grading}: Given the inconsistency of character damage degrees, as shown in Fig.~\ref{fig: grade}, we categorize the damage of characters into three levels: 
\begin{itemize}
    \item Severe damage: Characters exhibit complete loss of structural integrity, rendering them illegible even to expert examination.
    \item Medium damage: Characters show significant structural damage but remain identifiable through careful examination.
    \item Light damage: Characters maintain most structural features, enabling reliable identification despite visible damage.
\end{itemize}
\textbf{(3) Content Annotation}: We employ a differentiated approach for annotation depending on the condition of the characters. For light damage, direct visual annotation is performed. For medium damage, we attempt visual identification, and then verification using historical literature. For severe damage, annotation is conducted through the examination of multiple historical sources.
Our annotation processes are based on authoritative historical literature, such as \citet{cbeta}, \citet{nlc}, and \citet{jinshanzang}.
Through this process, we construct a comprehensive dataset that includes character-level and line-level bounding box annotations, character content labels, and damage grades. 
Notably, to ensure dataset quality, every image in our dataset was independently annotated by at least two experts through a rigorous validation process.
The entire manual workflow, including collection, annotation, and validation, requires approximately 2,400 person-hours.

\begin{figure}[t]
  \includegraphics[width=0.98\columnwidth]{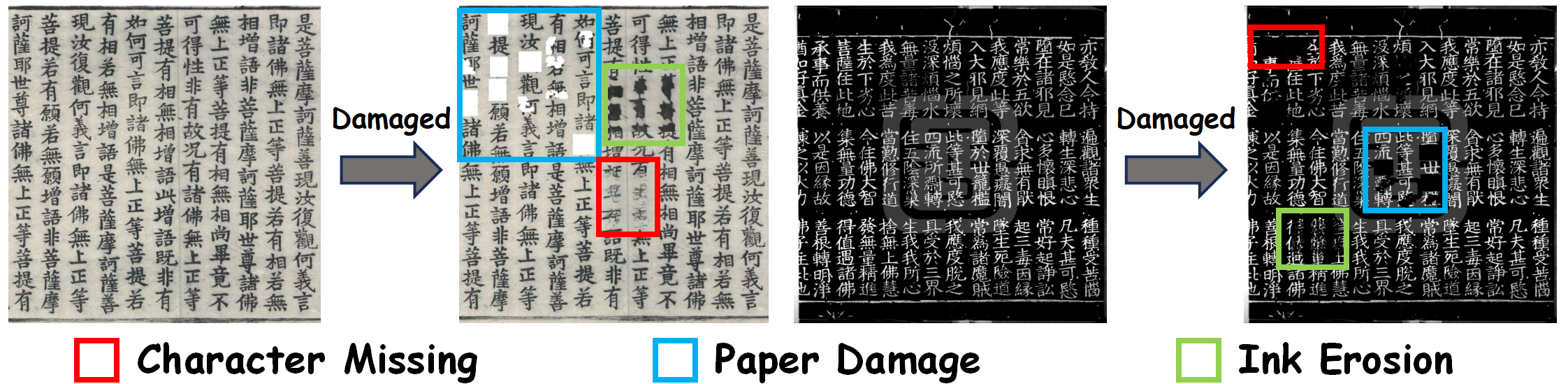}
  \centering

  \caption{Illustration of different damage types in the FPHDR dataset. Please zoom in for a better view.}
  \label{fig: syn}
  \vspace{-5mm}
\end{figure}

\begin{table}[t]
\resizebox{1\linewidth}{!}{
\begin{tabular}{lrrrrr}
\hline
Subset          & Images & Dam./im & Dam.   & Char/im & Char classes \\ \hline
Training set    & 6,543    & 51.40      & 293,195 & 452.37      & 15,208     \\
Test set     & 1,663    & 99.51      & 165,489 & 494.02      & 5,223      \\ \hline
\end{tabular}}

\caption{Statistics of the FPHDR dataset. "Dam." denotes the number of damaged characters.}
\label{tab: ana}
\vspace{-3mm}
\end{table}

\subsection{Damaged-Restored Pairs Data Synthesis}
\label{sec::syn}
As depicted in Sec.~\ref{sec: data collection}, we create synthetic training data by applying deterioration to well-preserved samples.
Based on the approach of \cite{diffhdr}, we construct pixel-level damaged-restored pairs samples comprising three types of deterioration, as listed in Fig.~\ref{fig: syn}:
(1) Character Missing: Content removal is performed using LAMA~\cite{suvorov2022resolution} on randomly generated masks.
(2) Paper Damage: Random areas in image patches are masked in black or white to simulate deterioration.
(3) Ink Erosion: Water erosion and fading effects are simulated by applying genalog's~\cite{genalog} diverse degradation modes and kernels.

\subsection{Dataset Analysis}
As shown in Tab.~\ref{tab: ana}, the statistical analysis of the data indicates that the average number of characters per sample is similar between the training and test sets. However, the training set includes a larger number of character categories, which helps the model learn more diverse character representations. In contrast, the average number of damaged characters per sample in the test set is higher, presenting a greater challenge to the restoration model's robustness. 
For more details, please refer to Appendix~\ref{appendix: dataset}.

\section{Methodology}

\subsection{Overall Framework}
The proposed AutoHDR architecture is illustrated in Fig.~\ref{fig: pipe}, containing three distinct yet interconnected stages: OCR-Assisted Damage Localization (OADL) for character recognition and damage localization, Damaged Content Prediction (DCP) for text restoration, and Historical Appearance Restoration (HAR) for pixel-level reconstruction.
The modular design enables independent training and execution while maintaining seamless integration.

\subsection{OCR-Assisted Damage Localization}
The OCR-assisted damage localization stage is primarily responsible for recognizing legible characters and detecting the locations of damaged characters.
To achieve this, we develop a character-level OCR model using data from various Chinese historical datasets, including MTHv2~\cite{mth}, M$^5$HisDoc~\cite{m5}, AHCDB~\cite{ahcdb}, and HisDoc1B~\cite{hisdoc1b}. This model demonstrates excellent performance on the test set, achieving a character localization F1 score of 98.35\% and a character recognition accuracy of 96.93\% under an Intersection over Union (IoU) threshold of 0.7. 
We then develop a model to localize severely damaged characters based on DINO~\cite{dino}.
After training the two models, we implement a localization fusion mechanism to merge the localization boxes from both models. Specifically, characters with an OCR confidence score below 0.1, indicating ambiguity, are designated as damaged, and their corresponding localization boxes $B_o$ are extracted. In parallel, we extract the localization boxes $B_s$ from the damage localization model. 
Then, we calculate the IoU between all $B_o$ and $B_s$. If $b_o \in B_o$ has an IoU greater than 0.5 with any $b_s \in B_s$, $b_o$ is removed. Conversely, if $b_o$ does not overlap with any $B_s$ (IoU lower than 0.5), $b_o$ is retained:
\begin{equation}
\label{eq: fusion}
\scalebox{0.95}{$B = B_s \cup \{b_o \in B_o | \max\limits_{b_s \in B_s} \text{IoU}(b_o, b_s) \leq 0.5\},$}
\end{equation}
We evaluate damage localization extensively in Sec.~\ref{sec: compare}, demonstrating its achieves human-comparable performance.
By arranging the character and damage bounding boxes in the natural reading order, we generate a sequence that specifies the positions of damaged characters, serving as input for the subsequent content prediction module.

\subsection{Damaged Content Prediction}
\label{sec: text pred}

Typically, historians first identify legible content from visual perception before restoring the incomplete or missing portions. Inspired by this paradigm, we combine both OCR's visual recognition and LLM's linguistic expertise to predict damaged content in the Damaged Content Prediction (DCP) stage.

We first adopt Qwen2 \cite{qwen2}, an advanced LLM, as our backbone model, specializing it in historical text prediction ability with a two-stage fine-tuning strategy.
In the first stage, inspired by~\citet{tonggu}, we conduct incremental pre-training using data from Daizhige~\cite{daizhige} and HisDoc1B (which encompass historical documents, poetry, art, Buddhist text, etc.) to enhance the model's comprehension of classical Chinese.
In the second stage, we fine-tune the model on pairwise damaged-restored historical texts from \citet{cbeta} (an authoritative Buddhist text repository) to enhance its content prediction ability.
We employ sequential mask tokens to represent damaged characters, directing the model to predict the corresponding contents.
The forms of the LLM's input and output are illustrated in Fig.~\ref{fig: pipe} (Stage 2).
A persisting issue is the inclusion of variant characters in classical Chinese texts, i.e., characters sharing identical meanings but differ in written form. Their rare occurrence challenges the model to recognize their equivalence to standard characters, hindering overall understanding.
To tackle this, we augment the data using character variants (detailed in Appendix~\ref{appendix: DCP}).
After training, the model acquires the capability to restore damaged content effectively.

While the trained LLM shows impressive content restoration performance, we discover that predicting the damaged content remains challenging due to the inherent complexity of classical Chinese, where multiple reasonable results could fit naturally in the same position. 
Therefore, relying solely on this LLM cannot guarantee the accuracy of text restoration. 
From the perspective of visual perception, we observe that OCR methods can recognize lightly damaged characters. This could serve as valuable auxiliary information to reduce the volume of damaged content requiring prediction and alleviate LLM's prediction burden. Motivated by this insight, we propose Vision-Language Context Prediction (VLCP), which leverages OCR for lightly damaged content recognition while allowing the LLM to focus on severely damaged content.

The procedure of VLCP is detailed in Algorithm~\ref{vlcp}. For each character, we first recognize its content through OCR. When OCR confidence exceeds a pre-defined threshold, we adopt its prediction directly. Otherwise, we score Top-k predictions from both OCR and LLM through the following strategies. For each candidate character (from the union of OCR and LLM prediction results), we compute a composite score incorporating:
\textbf{(1) Base Score}: A weighted sum of OCR and LLM probability scores. OCR achieves high confidence for lightly damaged characters but low confidence for severe damage, while LLMs excel in the latter case. Such complementarity allows our system to adaptively select predictions based on damage level.
\textbf{(2) Ranking Score}: A score is derived from characters' ranking positions in both models' predictions. Specifically, we rank the probabilities output by the LLM and OCR model separately, with each model generating its own ranking score based on the order of predictions according to their probabilities. This ranking criterion helps distinguish similar characters when their probability scores are close.
\textbf{(3) Matching Bonus}: Characters appearing in both models' predictions receive a bonus score, indicating visual and semantic plausibility.
Finally, we sum the above scores to obtain the composite score. 
The candidate character with the highest composite score is selected as the final prediction. 
At this point, the damaged content has been restored. 

\textbf{Discussion.} DCP stands as a crucial step to enable the full automation of the proposed AutoHDR. Since existing methods either necessitate manually inputting damaged content \cite{diffhdr} or retrieving text from a limited database \cite{zhumm}, the DCP firstly transcends these limitations by automatic prediction, achieving high restoration performance without human efforts. So far, we have obtained the coordinates of the damaged positions and their corresponding content, which will be used for the next stage.

\subsection{Historical Appearance Restoration}

Adhering to the "restoring the old as old" principle, we develop a diffusion model to restore the damaged historical appearance at the pixel level, built based on DiffHDR~\cite{diffhdr}, as depicted in Fig.~\ref{fig: pipe} (Stage3).
The model takes a damaged image $x_d$ as input and generates a restored image $x_r$ under the guidance of a mask image $x_m$ (indicating damaged regions) and a content image $x_c$ (specifying damaged content).
Specifically, we corrupt the $x_r$ by adding Gaussian noise to obtain the noised image $x_n$.
Then, the model input consists of four concatenated components:  $x_n$, $x_d$ $\in \mathbb{R}^{3\times H\times W}$, and $x_c$, $x_m$ $\in \mathbb{R}^{1\times H\times W}$. These form an 8-channel tensor that is processed by a denoiser $\mathcal{F}$ to generate the $x_r$.
The training objective of the model is as follows:
\begin{equation}
\mathcal{L}=\left\|\boldsymbol{x}_g-\mathcal{F}\left(\boldsymbol{x}_n ; \boldsymbol{x}_d, \boldsymbol{x}_c, \boldsymbol{x}_m\right)\right\|^2,
\end{equation}
where $x_g$ denotes the ground truth image.
After training, the model performs pixel-level restoration that maintains character style consistency and background feature similarity by leveraging the intact regions in damaged image $x_d$. 

\begin{algorithm}[t]
\caption{Vision-Language Context Prediction}
\fontsize{9.5}{11}\selectfont
\label{vlcp}
\begin{algorithmic}[1]
\Require{
    Input text $\mathcal{T}$;
    OCR model $\mathcal{O}$, Language model $\mathcal{L}$;
    OCR threshold $\tau$;
    OCR, LM weights $w_o, w_l$;
    Ranking score weight $\alpha$;
    Matching bonus $\beta$;
    TopK $k$
}
\Statex \textit{* $s_{ocr}$: OCR score; $s_c$: final candidate score}
\For{$p \in \mathcal{T}_{damaged}$}
    \State $s_{ocr} \gets \mathcal{O}(p)$
    \If{$s_{ocr}.conf > \tau$}
        \State $pred_p \gets s_{ocr}.pred$
    \Else
        \State $P_o \gets \mathcal{O}(p).topk$, $P_l \gets \mathcal{L}(p).topk$
        \For{$c \in P_o \cup P_l$}
             \State $r_o \gets \text{rank of } c \text{ in } P_o \text{, else } k$
            \State $r_l \gets \text{rank of } c \text{ in } P_l \text{, else } k$
            \State $s_c \gets w_o p_o + w_l p_l + \alpha(2k-r_o-r_l)$
            \State $s_c \gets s_c \cdot (\beta \text{ if } c \in P_o \cap P_l \text{ else } 1)$
        \EndFor
        \State $pred_p \gets \argmax_{c}(s_c)$
    \EndIf
\EndFor
\State \Return $pred$
\end{algorithmic}
\end{algorithm}

\begin{table*}[h]
\centering
\renewcommand{\arraystretch}{1.1}
\footnotesize
\resizebox{1\linewidth}{!}{
\begin{tabular}{lccccc}
\hline
\textbf{Method} & \textbf{Venue} & \textbf{Top1 w/o VLCP} & \textbf{Top1 w/ VLCP} & \textbf{Top5 w/ VLCP}  \\ 
\hline
SikuBERT~\cite{sikubert} & HuggingFace'22 & 40.49\% & 83.57\% (\textcolor{green}{+43.08\%}) & 87.28\% \\
Ithaca~\cite{Ithaca} & Nature'22 & 39.78\% & 86.73\% (\textcolor{green}{+46.95\%}) & 91.15\% \\
GujiBERT~\cite{gujibert} & arXiv'23 & 45.57\% & 83.58\% (\textcolor{green}{+38.01\%}) & 87.23\% \\
\hline
AutoHDR-MegatronBERT-1.3B & This work & 46.21\% & 83.42\% (\textcolor{green}{+37.21\%}) & 86.59\% \\
AutoHDR-Qwen2-1.5B & This work & \underline{50.49\%} & \underline{92.55\%} (\textcolor{green}{+42.06\%}) & \underline{96.83\%} \\
AutoHDR-Qwen2-7B & This work & \textbf{64.80\%} & \textbf{95.15\%} (\textcolor{green}{+30.35\%}) & \textbf{97.75\%} \\ 
\hline

 OCR-Only & This work & - & 82.13\% & - \\ 
\hline
\end{tabular}
}
\caption{Comparison of damaged content prediction results with existing methods. 
Our model variants are built upon Erlangshen-MegatronBERT\cite{fengshenbang} and Qwen2\cite{qwen2}.
}
\label{tab: dtp}
\vspace{-4mm}
\end{table*}

\begin{table}[h]
\centering
\resizebox{1\linewidth}{!}{
\begin{tabular}{lcccc}
\hline
\textbf{Method}     & \textbf{Precision} & \textbf{Recall} & \textbf{F1 score} \\ \hline
YOLOv7~\cite{yolov7}      & 87.1               & 86.4            & 86.5              \\
Co-DETR~\cite{codetr}     & 80.8               & 87.4            & 83.7              \\
DINO~\cite{dino}          & 97.0               & 91.4            & 94.1              \\ \hline
Historian*         & 98.9               & 95.6            & 97.2              \\ \hline
\end{tabular}
}
\caption{Comparison of damage localization results across different methods. * indicates that only a subset of the data is evaluated.}
\label{tab: det}
\end{table}

While this model performs well, it is limited to patch-level restoration. To extend it to page-level, we introduce a Patch-AutoRegressive (PAR) mechanism during inference.
PAR begins by dynamically selecting the starting patch from the four corners of the damaged image, choosing the one with the least number of damaged characters to ensure the model has the most intact characters for reference.
The selected patch is restored and placed back in its original location.
Then, an overlap sliding window operation extracts the next patch, leveraging previously restored regions as references for further restoration. 
To avoid split characters caused by the sliding window, we apply a mask to these regions during processing, ensuring all restored characters are complete. 
The process iterates until full-page restoration is complete. By leveraging references from previously restored patches, the PAR ensures visual consistency across the full page.

PAR exhibits significant practical value in engineering applications by addressing common challenges in HDR, such as the limitation to patch-level restoration and the difficulty in maintaining consistency across the entire page. We demonstrate its effectiveness in Sec.~\ref{sec: pam}, with detailed pseudocode provided in Appendix (Algorithm~\ref{alg: dynpatch}).

\begin{table}[t]
\centering
\resizebox{0.7\linewidth}{!}{
\begin{tabular}{lc}
\hline
\textbf{Method} & \textbf{Accuracy} \\ 
\hline
Historian         & 44.08\% \\ 
AutoHDR-Qwen2-7B              & 76.38\% \\
Historian + AutoHDR & 85.05\% \\
\hline
\end{tabular}
}
\caption{Evaluating AutoHDR's collaboration.}
\vspace{-3mm}
\label{tab:human-eval}
\end{table}

\begin{table*}[t]
\centering
\resizebox{0.98\linewidth}{!}{
\begin{tabular}{lccccccc}
\hline
\multirow{2}{*}{\textbf{Method}}  & \multicolumn{3}{c}{\textbf{AR (\%)}} &   &  \multicolumn{2}{c}{\textbf{User Study} ↑}  & \multirow{2}{*}{\textbf{LPIPS} ↓}  \\ \cline{2-4}  \cline{6-7} 
             & \textbf{Light} & \textbf{Medium} & \textbf{Severe} &  & \textbf{Style Consistency} & \textbf{Overall Quality (\%)} \\ \hline
Damaged Documents & 77.42  & 68.98 & 46.83 &  & - & 0.00  & -\\
NAFNet~\cite{nafnet}      & 78.56              & 73.56              & 61.07 &  & 2.721   & 2.57   & \underline{0.0585}\\
Uformer~\cite{Uformer}       & 78.03         & 72.72         & 62.94 & & \underline{3.153}  & \underline{8.75}  & 0.0633 \\
Restormer~\cite{Restormer}       & 87.27          & 84.40      & 75.98 & & 2.877 & 3.90  & 0.0691\\
\hline
AutoHDR (Ours)      &  \underline{91.80}    &  \underline{90.01}     &  \underline{84.05} & & \textbf{3.934} & \textbf{84.78}  & \textbf{0.0541} \\ 
Historian + AutoHDR (Ours)  & \textbf{93.63}   & \textbf{93.81}      & \textbf{94.25}   & & -  & - & - \\ \hline
\end{tabular}
}
\caption{Comparison of historical appearance restoration results with existing methods.}
\label{tab: hap}
\end{table*}

\section{Experiments}
\label{sec: exp}

\subsection{Evaluation Metrics}

For damage localization, performance is evaluated using the F1 score, precision, and recall at an IoU threshold of 0.5. 
For damaged content prediction, Top-1 and Top-5 accuracy metrics are adopted. 
For appearance restoration, since obtaining pixel-level ground truth from real data is extremely difficult, we evaluate the restoration quality through character recognition accuracy. Specifically, we train a text-line OCR using AHCDB, MTHv2, and M$^5$HisDoc to recognize the restored data, and adopt the commonly used Accurate Rate (AR)~\cite{hiercode} as our evaluation metric. The formula for AR is as follows:
\begin{equation}
\scalebox{0.90}{$AR = (N_{t} - D_{e} - S_{e} - I_{e})/N_{t},$}
\end{equation}
where $N_t$ is the total number of characters in annotations, while $D_e$, $S_e$, and $I_e$ denote deletion, substitution, and insertion errors, respectively.
For pixel-level evaluation on synthetic data, we use LPIPS~\cite{lpips} as the evaluation metric, since \citet{diffhdr} demonstrated that PSNR and SSIM are unsuitable for historical document restoration tasks.

\begin{figure*}[t]
  \includegraphics[width=1\linewidth]{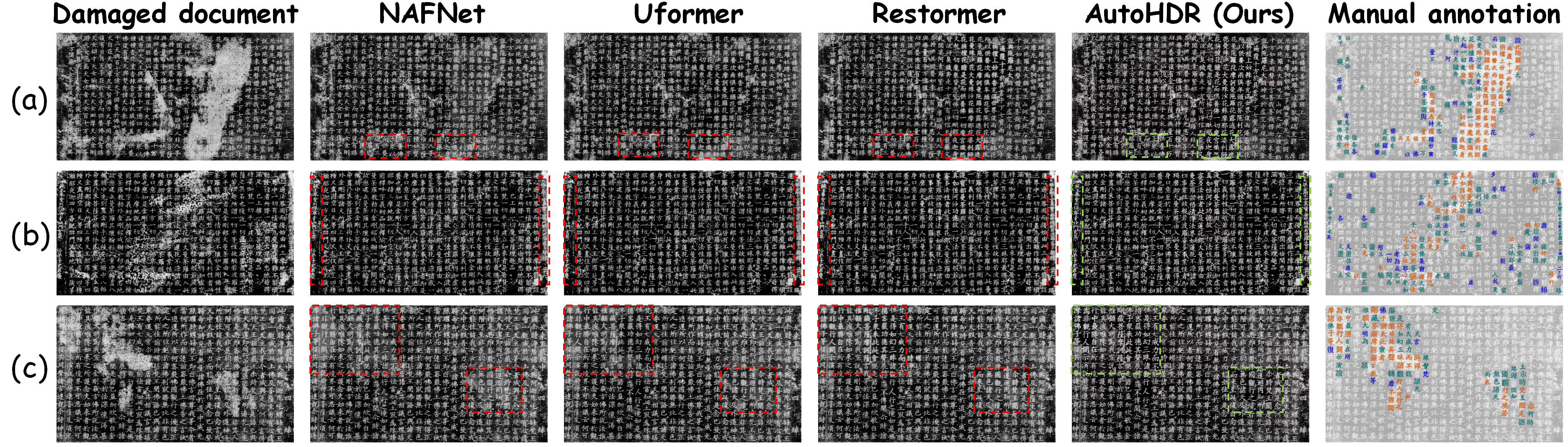}
  \centering
  \vspace{-4mm}
  \caption{Qualitative comparison. We visualize the results of some evaluated methods. Red highlights regions with varying degrees of restoration inaccuracies, while green denotes areas with satisfactory restoration quality.}
  \label{fig: compare}
\end{figure*}

\subsection{Comparison with Existing Method}
\label{sec: compare}

\textbf{Damage Localization:}
We train DINO~\cite{dino}, Co-DETR~\cite{codetr}, and YOLOv7~\cite{yolov7} on the FPHDR dataset. As shown in Tab.~\ref{tab: det}, DINO achieves the best performance with an F1 score of 94.1\%. Additionally, we invite historians to evaluate a randomly selected set of 30 samples, achieving an F1 score of 97.2\%.
These historians (external to our annotation team) were not familiar with our strict annotation process, and for some characters with light damage that were still recognizable, they deemed restoration unnecessary, leading to a less than 100\% F1 score in human evaluation.
Overall, using DINO as the localization model is already comparable to human performance. 
In addition, when working collaboratively with historians, it can provide high-quality initial localization, allowing historians to make minor adjustments to achieve better detection results, thereby significantly reducing manual workload.

\noindent\textbf{Damaged Content Prediction:}
We compare our method with SikuBERT~\cite{sikubert}, GujiBERT~\cite{gujibert}, and Ithaca~\cite{Ithaca}. For a fair comparison, we retrain them following the approach in Sec.\ref{sec: text pred}. 
As shown in Tab.~\ref{tab: dtp}, AutoHDR models outperform other methods in both Top-1 and Top-5 accuracy, with larger models achieving better performance. Notably, the VLCP significantly improves Top-1 accuracy across all methods by an average of 39.61\%. Moreover, as shown in the fourth column, all models outperform the OCR-Only baseline after incorporating VLCP. These results show that VLCP enables the model to classify damage grades automatically, using OCR for recognizable characters and LLM for unrecognizable ones, highlighting the effectiveness of the proposed VLCP.

Furthermore, our method achieves a maximum Top-5 accuracy of 97.75\%, demonstrating that AutoHDR can provide valuable suggestions for historians. To validate its collaborative potential, we test 23 severely damaged documents in three scenarios (Tab.~\ref{tab:human-eval}): historian-only (44.08\%), AutoHDR-only (76.38\%), and collaborative predictions where historians select from AutoHDR's Top-5 suggestions (85.05\%). These results highlight AutoHDR's collaborative capability, offering critical support for restoring and studying historical texts.

\begin{figure}[t]
  \includegraphics[width=0.98\linewidth]{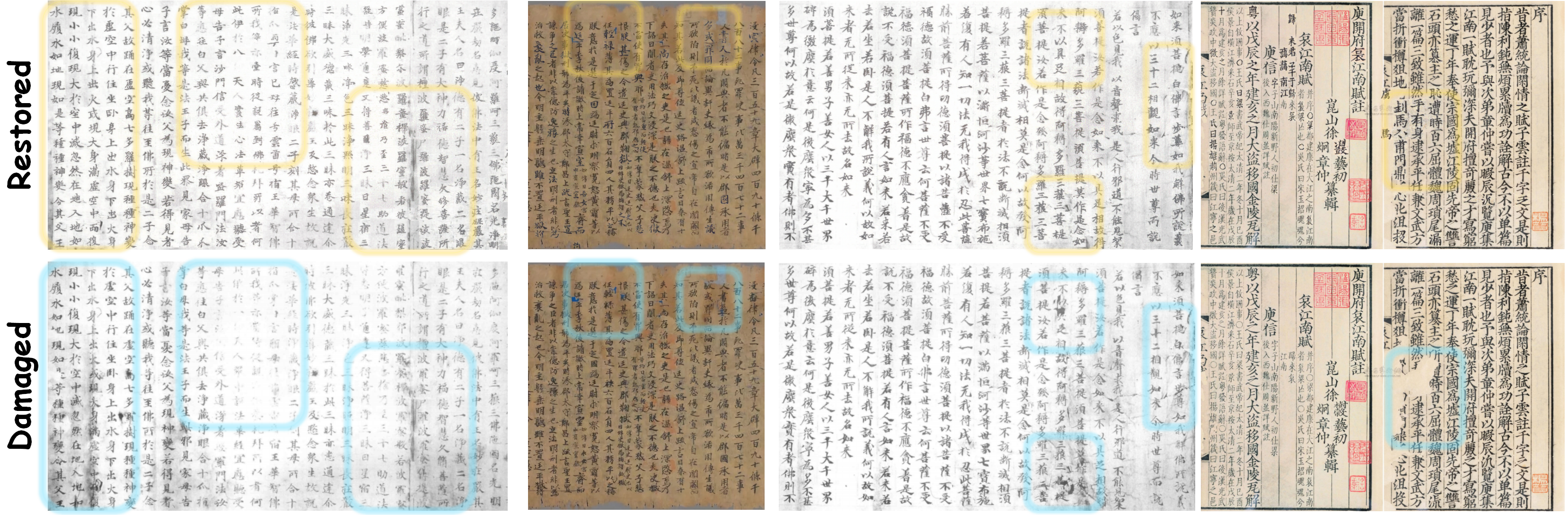}
  \centering
  \caption{Restoring different types of documents.}
  \label{fig: diff}
  \vspace{-4mm}
\end{figure}

\noindent\textbf{Historical Appearance Restoration:}
We compare our method with three state-of-the-art methods: NAFNet~\cite{nafnet}, Uformer~\cite{Uformer}, and Restormer~\cite{Restormer}. 
All methods are trained using the same procedure as DiffHDR~\cite{diffhdr}. They receive identical input from the first two stages of AutoHDR and are required to output the restored images.
As shown in Tab.~\ref{tab: hap}, AutoHDR achieves SOTA performance. Compared to the original damaged images, it improves the recognition accuracy by 14.38\%, 21.03\%, and 37.22\% on light, medium, and severe damage grades, respectively, demonstrating the strong restoration capability of our solution.
Due to the difficulty of obtaining pixel-level annotations for real images, we conduct two user studies (Style Consistency and Overall Quality) with 20 participants to evaluate restoration quality.
For style consistency, participants are asked to score the font style similarity between restored and original regions on a 1-5 scale (5 = completely consistent, 1 = completely inconsistent), focusing solely on style while ignoring other factors like image clarity.
For overall quality, participants are asked to consider font style similarity, background integration, and character accuracy, then select the best result from the above four models.
As presented in Tab.~\ref{tab: hap}, our method achieves the highest score in the user study, indicating its superior capability in faithfully restoring the original appearance of historical documents.
Additionally, we select 100 intact images of the Fangshan Stone Sutra and degrade them according to the method described in Sec.~\ref {sec::syn} to evaluate pixel-level restoration performance. As shown in the last column of Tab.~\ref{tab: hap}, our model achieves the best performance with the lowest LPIPS.
Furthermore, we invite historians to collaborate with AutoHDR by reviewing and modifying the intermediate results at each stage of the process. As shown in the last row of Tab.~\ref{tab: hap}, this collaboration significantly improves performance, particularly on severe damage grade, achieving a 10.20\% improvement over AutoHDR only. This further underscores AutoHDR’s strong collaborative capability.

The qualitative results are visualized in Fig.~\ref{fig: compare}. NAFNet often shows character distortion and stroke loss, while Uformer and Restormer produce blurry regions in restored areas (see Fig.~\ref{fig: compare}(a)(c)). They also struggle with small or complex characters (see Fig.~\ref{fig: compare}(b)). In contrast, AutoHDR achieves superior performance. Additionally, Fig.~\ref{fig: diff} illustrates AutoHDR's generalization by restoring various historical documents.

The outstanding performance of AutoHDR, coupled with its fully automated capabilities, underscores its practicality and potential for widespread application. Collaborative efforts with historians further reinforce the effectiveness and utility of AutoHDR, making it a valuable tool for the restoration and study of historical documents.

\begin{figure}[t]
  \includegraphics[width=0.98\linewidth]{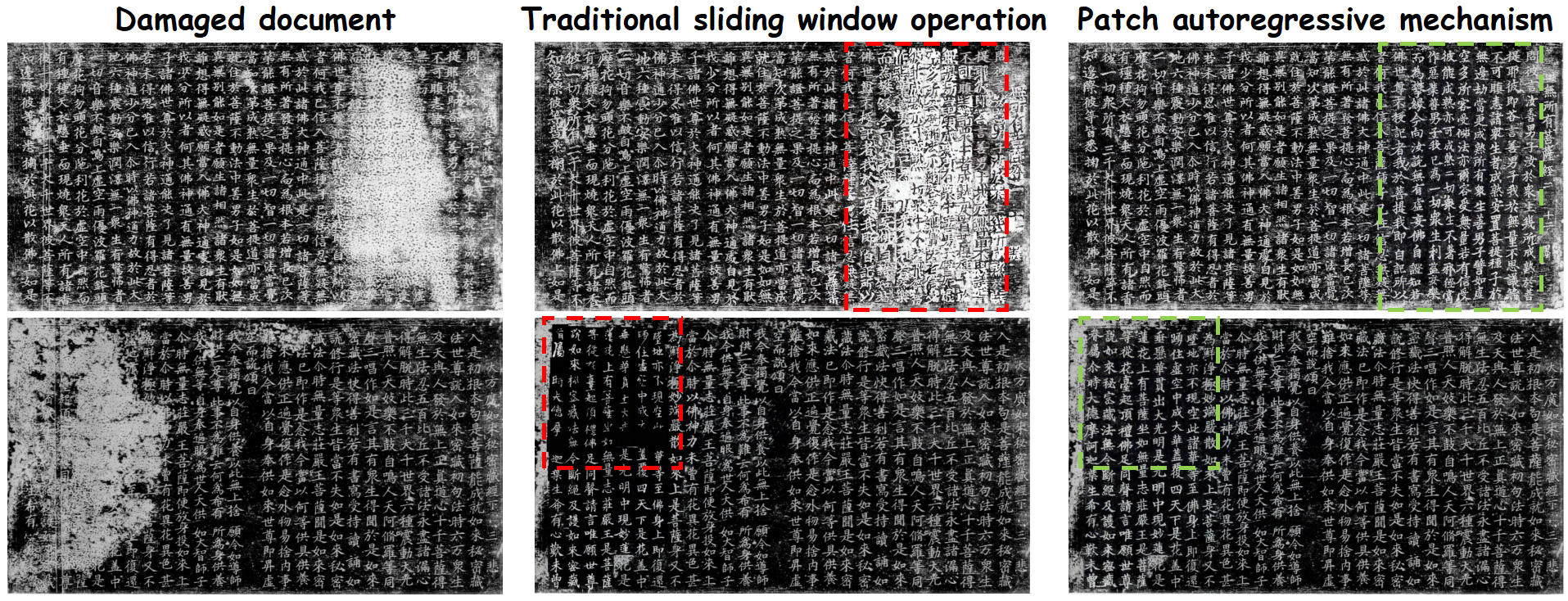}
  \centering
  \caption{Effectiveness of patch autoregressive. }
  \label{fig: dya}
  \vspace{-5mm}
\end{figure}

\subsection{Ablation Study}
\label{sec: pam}
To validate the effectiveness of our proposed patch-autoregressive mechanism, we compared it with traditional sliding window operations. As shown in Fig.~\ref{fig: dya}, the traditional sliding window operation may lead to restoration failure or incomplete character restoration. Conversely, our patch-autoregressive mechanism achieves high-quality page-level restoration.
Additionally, we conduct ablation studies on the LLM's input/output formats, the data augmentation method, and the VLCP algorithm in Appendix~\ref{appendix: abla}. These studies demonstrate that our current format design and proposed methods are effective.

\section{Conclusion}

In this paper, we propose AutoHDR, a novel solution for HDR that mimics historians' restoration practices through a three-stage approach: OCR-assisted damage localization, vision-language context text prediction, and patch autoregressive appearance restoration. 
AutoHDR's modular architecture enables seamless collaboration between AI and historians, allowing for flexible intervention and enhancement at various stages of the restoration process. 
Furthermore, we present FPHDR, a pioneering full-page HDR dataset containing 6,543 synthetic samples for training and 1,633 annotated real samples for evaluation. 
Extensive experimental results demonstrate AutoHDR's outstanding performance in HDR tasks and its effectiveness in supporting historians' work. 
We anticipate that this research will significantly advance AI-assisted HDR and make a substantial contribution to cultural heritage preservation.

\section*{Acknowledgements}

This research is supported in part by the National Natural Science Foundation of China (Grant No.: 62476093, 62441604).

\section*{Limitations}

AutoHDR utilizes a three-stage process for the restoration of historical documents, which inherently introduces certain limitations in processing speed. In our experiments, inference on a single NVIDIA A10 GPU requires an average of approximately five minutes per image. Furthermore, as indicated in Tab.~\ref{tab: hap}, although our method has achieved promising performance, the restoration results may still exhibit inaccuracies, particularly in scenarios involving severe document degradation. Therefore, collaboration with historians emerges as a more robust and reliable strategy for document restoration and research. In future work, we will explore the feasibility of utilizing large vision-language models (such as Qwen2.5-VL~\cite{qwen2.5-VL} and InternVL 2.5~\cite{InternVL2.5}) to perform end-to-end restoration of historical documents.

\section*{Ethical Statements}

This research is dedicated to advancing historical document restoration, ensuring positive contributions to cultural preservation. While AutoHDR offers significant advantages in restoring damaged documents, we remain mindful of the potential risks of misuse, such as the generation of falsified historical records. To address these risks, we apply strict licensing agreements that limit the dataset and code to academic research and non-commercial use, ensuring the technology is applied ethically and responsibly.

\bibliography{acl_latex}

\clearpage
\appendix

\section{Dataset Details}
\label{appendix: dataset}
\begin{figure*}[hb]
  \includegraphics[width=1\linewidth]{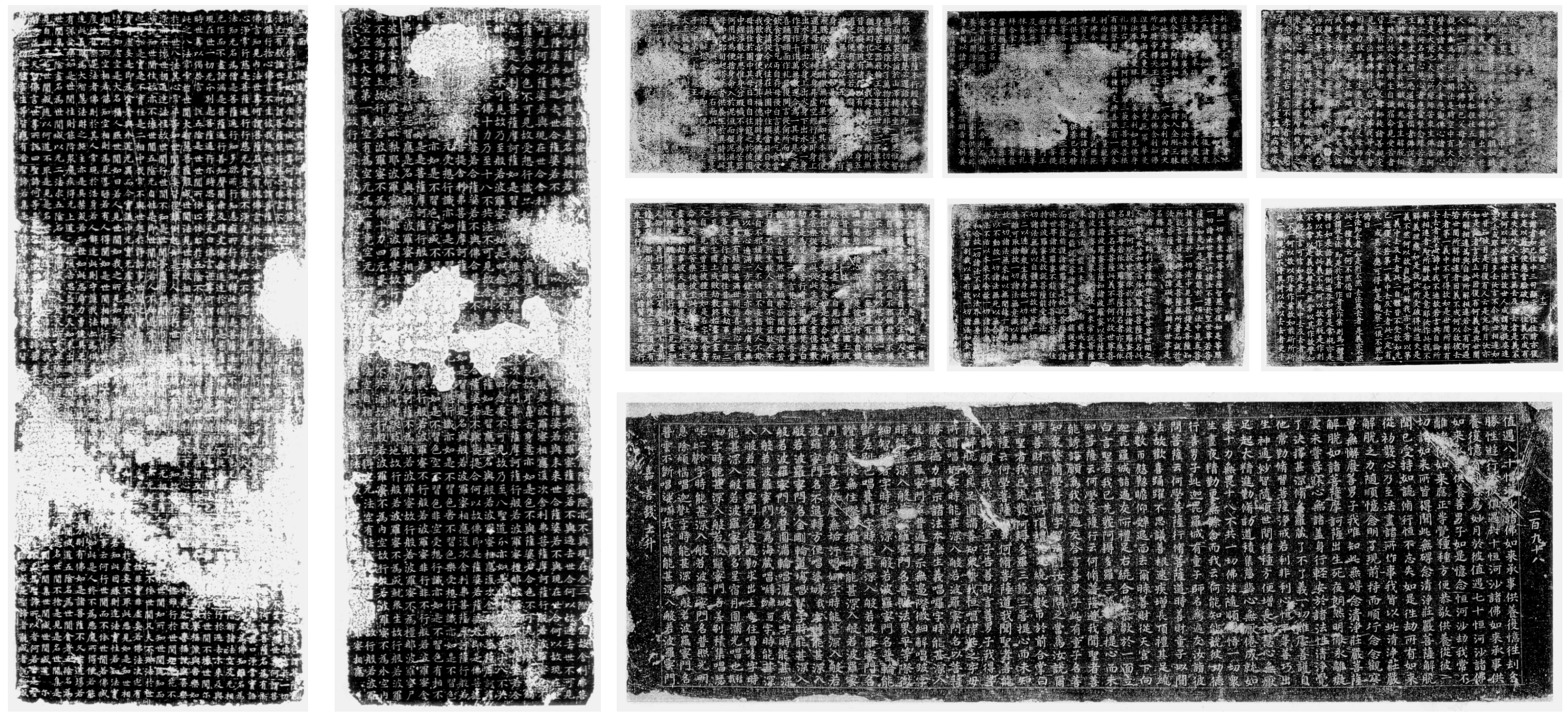}
  \centering
  \vspace{-3mm}
  \caption{Examples of real samples in the FPHDR dataset. }
  \label{fig: real sample}
  \vspace{-3mm}
\end{figure*}

\begin{figure*}[h]
  \includegraphics[width=1\linewidth]{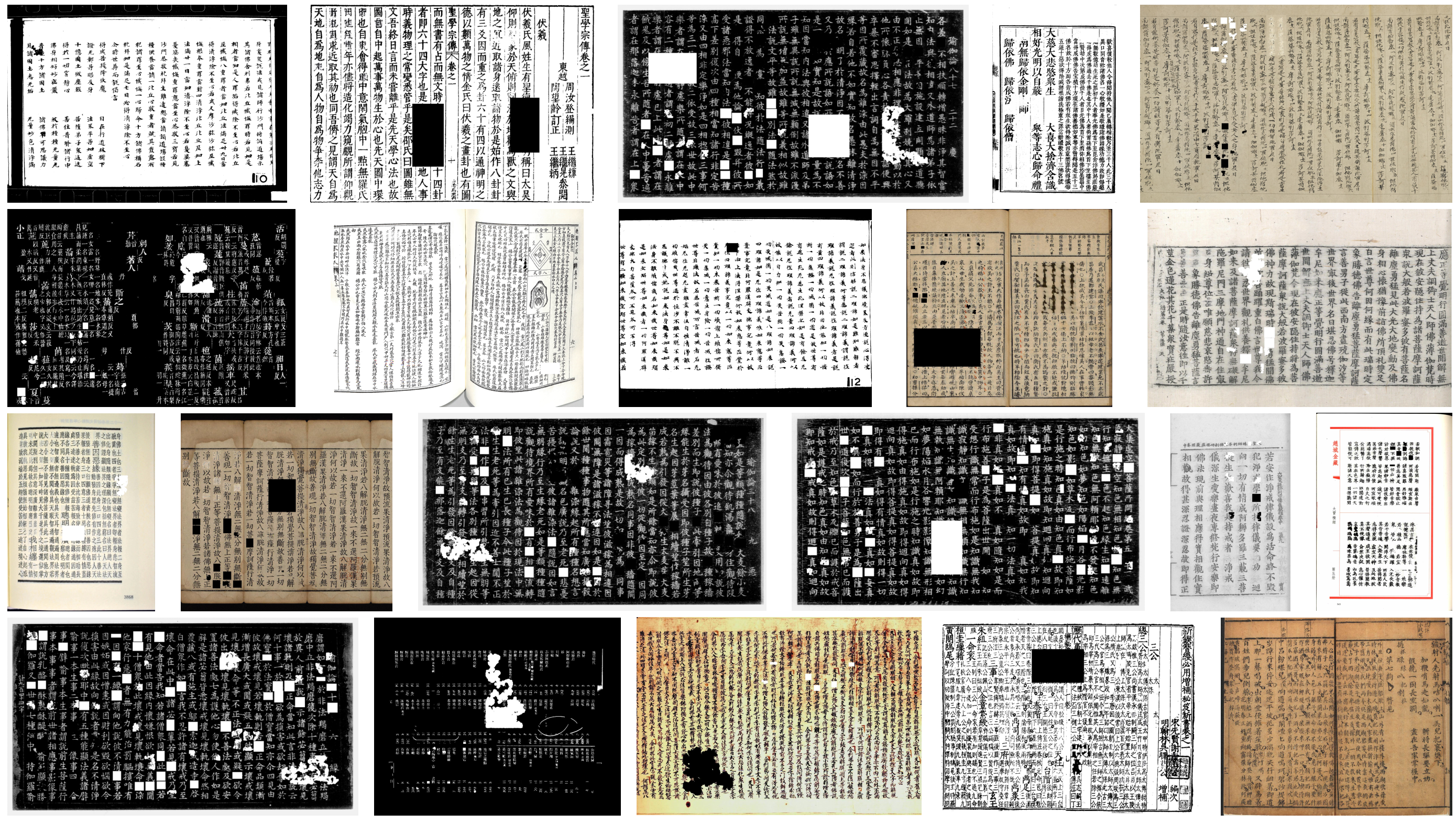}
  \centering
  \vspace{-3mm}
  \caption{Examples of synthetic samples in the FPHDR dataset. }
  \label{fig: syn sample}
  \vspace{-3mm}
\end{figure*}

In this section, we present the details of our dataset. 
Tab. \ref{tab:test_stats} shows the distribution of degradation grades (light, medium, and severe) in the FPHDR test set, which contains 1,663 images with a total of 165,489 degraded characters. Furthermore, we provide additional visualizations from our FPHDR dataset, including real samples in Fig.~\ref{fig: real sample} and synthetic samples in Fig.~\ref{fig: syn sample}.

\section{Implementation Details}

\subsection{Character-Level OCR Model}

The OCR model used in the OCR-Assisted Damage Localization stage first utilizes YOLOv7~\cite{yolov7} for character detection, followed by ViT-Base~\cite{vit} for character recognition. This model is trained on the MTHv2~\cite{mth}, M$^5$HisDoc~\cite{m5}, AHCDB~\cite{ahcdb}, and HisDoc1B~\cite{hisdoc1b} datasets, with the division of training and testing data strictly following the official splits of these datasets.

\subsection{Damage Localization}

We implement and train two localization models (DINO\footnote{\url{https://github.com/open-mmlab/mmdetection/blob/main/configs/dino/dino-5scale_swin-l_8xb2-12e_coco.py}} and Co-DETR\footnote{\url{https://github.com/Sense-X/Co-DETR/blob/main/projects/configs/co_dino/co_dino_5scale_swin_large_1x_coco.py}}) based on the framework of MMDetection~\cite{mmdetection}. For both models, we employ SwinTransformer-Large~\cite{Swin} as the backbone, and all other configurations follow the default settings of MMDetection. For YOLOv7~\cite{yolov7}, we use the official source code\footnote{\url{https://github.com/WongKinYiu/yolov7}} for implementation. All models are trained using the pre-trained weights provided by the official sources. The training is conducted on 6 NVIDIA A800 GPUs. The image size is $1333 \times 1333$.
During training, we first pre-train the model using synthetic data, then randomly select 1,163 real images for fine-tuning, and evaluate the model on the remaining 500 real images.

\subsection{Damaged Content Prediction}
\label{appendix: DCP}

The hyper-parameter settings of incremental pre-training and content prediction fine-tuning are shown in Tab.~\ref{tab: hype}. All experiments are completed on 8 NVIDIA A800 GPUs. 

During content prediction fine-tuning, we simulate damaged texts using sequential mask tokens ([mask1], [mask2]...) to randomly replace characters, with masking ratios varying from 5\% to 90\%. 
To address the challenge caused by variant characters in classical Chinese texts, we propose a Variant-based Data Augmentation method (VDA). 
Specifically, we compile a reference table of 32,260 variant characters and randomly replace standard characters with their variants during data construction to improve the model's comprehension of variant characters.
Additionally, to enhance the model's robustness, we randomly remove characters with a 3\% probability during training. 

In the Vision-Language Context Prediction (VLCP) algorithm, we set the OCR threshold $\tau$ to 0.9, OCR and LM weights ($w_o$, $w_l$) to 0.6 and 0.4 respectively, Ranking score weight $\alpha$ to 0.05, Matching bonus $\beta$ to 1.5, and TopK $k$ to 5. These values have proven robust across a wide range of documents and damage conditions in our experiments. However, users can adjust these parameters based on the degree of document damage. For instance, for severely damaged documents, higher weights should be assigned to the language model to leverage contextual information, while for less damaged documents, higher weights can be assigned to the OCR model to prioritize visual recognition accuracy.

\subsection{Historical Appearance Restoration}
We train the appearance restoration model with a batch size of 16 and a total epoch of 195 and adopt an AdamW optimizer with $\beta_1 = 0.95$ and $\beta_2 = 0.999$. The learning rate is set as $1 \times 10^4$ with the linear schedule. The image size is $512 \times 512$. The training is conducted on 4 NVIDIA A6000 GPUs. Additionally, we adopt the DPM-Solver++ as our sampler with the inference step of 20.

The detailed procedure of the Patch Autoregressive mechanism (PAR) is presented in Algorithm.~\ref{alg: dynpatch}. For our PAR implementation, we configure the patch size $P$ to 448 and the stride $S$ to 224.

\begin{table}[t]
\centering
\resizebox{1\linewidth}{!}{
\begin{tabular}{lccccc}
\hline
\multirow{2}{*}{Subset} & \multirow{2}{*}{Images} & \multicolumn{3}{c}{Damage} & \multirow{2}{*}{Total} \\
\cline{3-5}
& & Light & Medium & Severe & \\
\hline
Test set & 1,663 & 27,231 & 96,114 & 42,144 & 165,489 \\
\hline
\end{tabular}
}
\vspace{-3mm}
\caption{The distribution of different damaged grades in the FPHDR test set.}
\vspace{-6mm}
\label{tab:test_stats}
\end{table}

\begin{table}[ht]
\centering
\resizebox{1\linewidth}{!}{
\begin{tabular}{ccc}
\hline

Hyperparameter & \makecell{Incremental\\Pretraining} & \makecell{Content prediction\\fine-tuning}\\
\hline
Precision & bf16 & bf16\\
Epoch & 1 & 5\\
Batch size & 288 & 240\\
Learning rate & 1e-5 & 6e-6\\
Weight decay & 0 & 0\\
Warmup ratio & 0.03 & 0.03\\
LR scheduler type & cosine & cosine\\
Optimizer & AdamW & AdamW\\
$\beta_1$ & 0.9 & 0.9\\
$\beta_2$ & 0.999 & 0.999\\
Max length & 3072 & 3072\\
\hline
\end{tabular}
}
\caption{Hyper-parameter settings in incremental pre-training and content prediction fine-tuning.}
\label{tab: hype}
\end{table}

\section{Ablation Study}
\label{appendix: abla}
\begin{figure}[t]
  \includegraphics[width=1\linewidth]{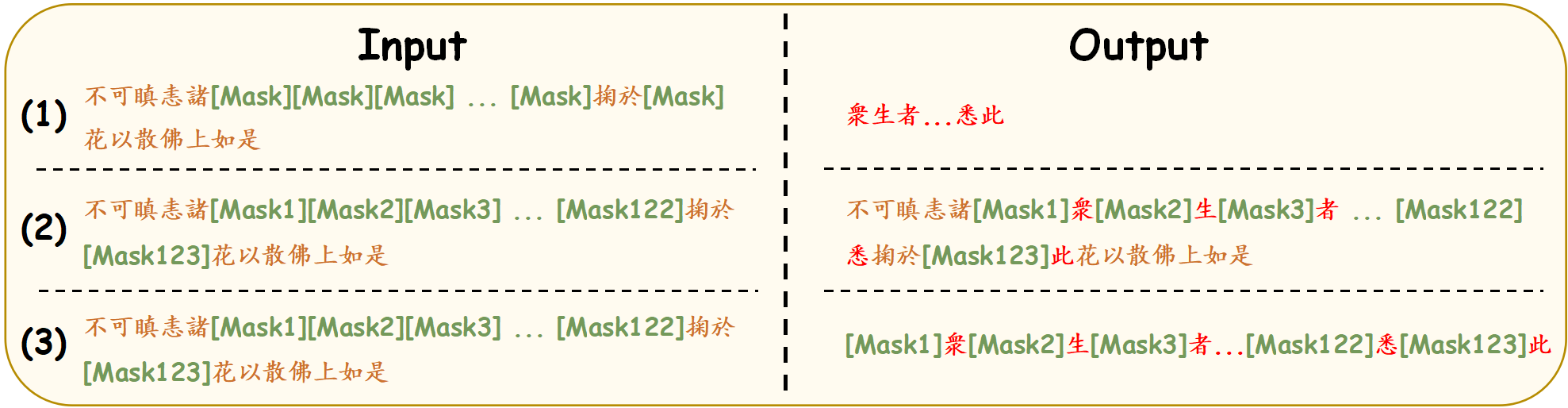}
  \centering
  \vspace{-5mm}
  \caption{Examples of input and output formats. }
  \label{fig: io}
  \vspace{-3mm}
\end{figure}

The ablation study is designed to investigate how different input-output formats affect the performance of text restoration and validate the effectiveness of our proposed Variant-based Data Augmentation (VDA) method.
As shown in Fig.~\ref{fig: io}, we design three types of input-output formats for damaged content prediction.  Format (1) uses a single mask token to represent damaged characters, outputting the prediction sequentially. Format (2) uses sequential mask tokens to represent damaged characters and outputs the restored text with mask tokens indicating damaged positions. Format (3) is the same as introduced in Sec.~\ref{sec: text pred}. 
Then, we conduct experiments using AutoHDR-Qwen2-1.5B~.
The experimental results in Tab.~\ref{tab: abla} indicate that formats (2) and (3) achieve comparable and better performance. 
However, format (3) provides a shorter output sequence, thus leading to faster inference speed, making it the preferred choice.
Furthermore, as shown in the last two columns of Tab.~\ref{tab: abla}, the proposed variant-based data augmentation method and VLCP demonstrate significant effectiveness.

\begin{table}[t]
\resizebox{1.0\linewidth}{!}{
\begin{tabular}{cccccc}
\hline
\multirow{2}{*}{{Method}} & \multicolumn{3}{c}{{Input/Output formats}} & \multirow{2}{*}{{+VDA}} & \multirow{2}{*}{{+VLCP}} \\ \cline{2-4}
                        & (1)          & (2)         & (3)         &                       &                        \\ \hline
Top1 Acc                & 35.72\%      & 40.43\%     & 40.32\%     & 50.49\%               & 92.55\%                \\ \hline
\end{tabular}
}
\vspace{-2mm}
\caption{Ablation study on input/output formats and Variant-based Data Augmentation (VDA) method (with VDA and VLCP based on format 3).}
\vspace{-4mm}
\label{tab: abla}
\end{table}

\section{More Visualization Results}
\begin{figure*}[h]
  \includegraphics[width=1\linewidth]{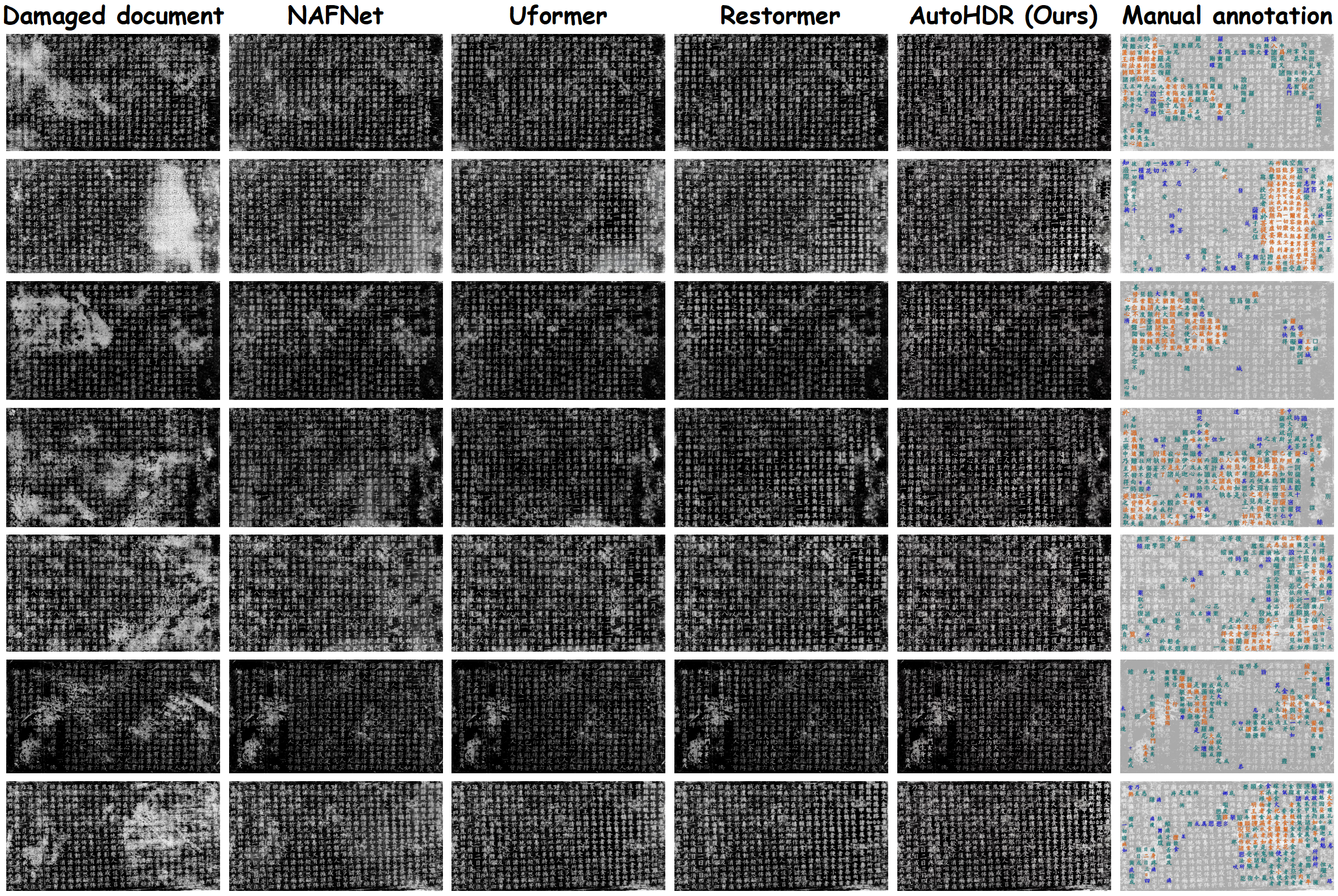}
  \centering
  \vspace{-3mm}
  \caption{Additional qualitative comparison. }
  \label{fig: add_compare}
  \vspace{-3mm}
\end{figure*}

As shown in Fig.~\ref{fig: add_compare}, we provide more visualization of restoration results from Restormer~\cite{Restormer}, NAFNet~\cite{nafnet}, Uformer~\cite{Uformer}, and AutoHDR. The visual comparison demonstrates that AutoHDR achieves superior performance.

\begin{figure*}[h]
  \includegraphics[width=1\linewidth]{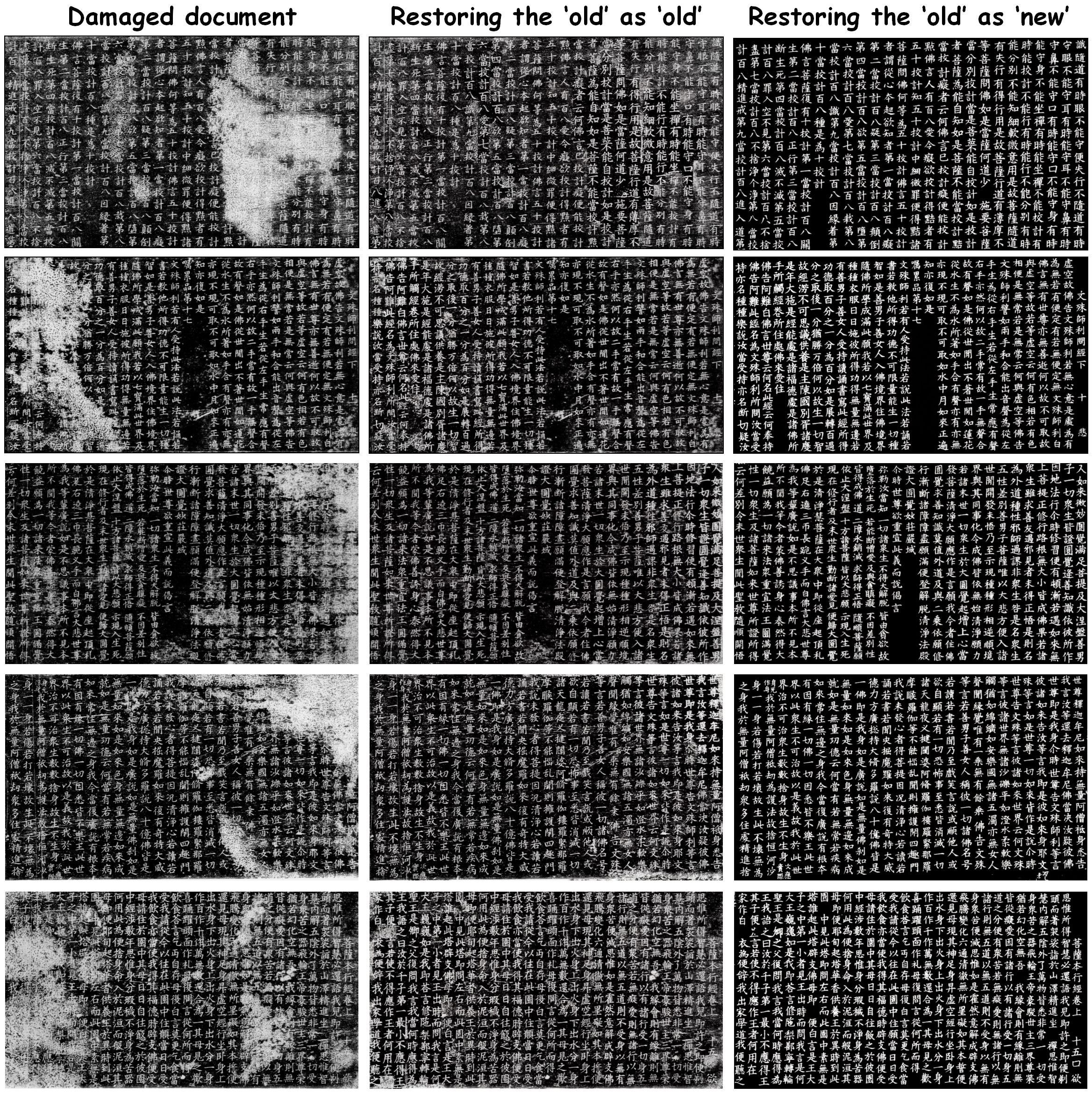}
  \centering
  \vspace{-3mm}
  \caption{Additional restoration results of AutoHDR.}
  \label{fig: old_new_ori}
  \vspace{-3mm}
\end{figure*}

Furthermore, we present additional restoration results of AutoHDR in Fig.~\ref{fig: old_new_ori}, which demonstrate its dual restoration capabilities. On the one hand, AutoHDR can effectively adhere to the principle of "restoring the old as old", maintaining font style consistency and background feature similarity. On the other hand, it can extend to "restoring the old as new", thus accommodating diverse user requirements for both heritage preservation and modern restoration.

\section{The Impact of Patch Size in Patch Autoregressive Mechanism}
Based on our observations, the optimal patch size should be determined relative to the character size in the document being restored. Specifically, the length and width of the patch should accommodate at least three characters (to ensure there are enough intact characters for reference). When this requirement is met, the patch size has minimal impact on page-level restoration. Typically, our default setting (patch size = 448) is sufficient to meet the needs of most practical applications.

\section{Potential Risks and Human-AI Collaboration Solutions}
Although the LLM is fine-tuned with historical corpora, and our model significantly improves the accuracy of damaged content predictions by combining OCR visual information with the semantic understanding of the LLM, there remain certain special cases where the model may still generate plausible but incorrect results.

Therefore, we recommend that the optimal use case for our model is in collaboration with historians.
Through extensive experimental validation (Tab.~\ref{tab: det}, Tab.~\ref{tab:human-eval}, and Tab.~\ref{tab: hap}), we found that allowing historians to review and modify the intermediate results of our model significantly enhances the accuracy and reliability of historical document restoration. This collaborative approach not only addresses the model's potential errors in ambiguous cases but also leverages domain expertise to ensure historical accuracy.

\begin{algorithm*}[t]
\caption{Patch-Autoregressive Mechanism}
\label{alg: dynpatch}
\begin{algorithmic}[1]
\Require{
    Damaged image $X_d$ with damage detection boxes $B=\{b_1, b_2, \dots \}$, 
    Patch size $P$, Stride $S$
}
\Ensure{
    Restored image $X_r$
}
\State $X_r \gets \text{Copy}(X_d)$
\State Initialize each $b_i \in B$ with $\mathrm{restored\_flag}(b_i) \leftarrow \text{False}$ 
\While{there exists an unrestored box $b_i \in B$ with $\mathrm{restored\_flag}(b_i) = \text{False}$}
    \State $\mathcal{U} \gets \{\,b_i \mid \mathrm{restored\_flag}(b_i)=\text{False}\}$ \Comment{Collect unrestored boxes}
    \State $(x_\mathrm{min}, y_\mathrm{min}, x_\mathrm{max}, y_\mathrm{max}) \gets \text{ComputeExtent}(\mathcal{U})$
    \State $\mathcal{C} \gets \text{DefineCorners}(x_\mathrm{min},y_\mathrm{min},x_\mathrm{max},y_\mathrm{max})$ 
       \Comment{Four corners for patch placement}
    \For{corner $c \in \mathcal{C}$}
        \State $\mathrm{cnt}(c) \gets \text{CountUnrestoredInPatch}(c, \mathcal{U}, P)$
        \Comment{Compute the number of unrestored boxes in the patch at corner $c$}
    \EndFor
    \State $c^* \gets \arg\min_{c \in \mathcal{C}} \bigl(\mathrm{cnt}(c)\bigr)$ 
       \Comment{Select corner with minimal damage}
    \For{(startX, startY) in SlidingWindow($c^*, S, P$)}
        \State $(x_s, y_s, x_e, y_e) \gets \text{ClipToBounds}(startX, startY, X_r, P)$
        \State $B_\mathrm{inside} \gets \text{FindFullyContainedBoxes}(x_s,y_s,x_e,y_e,\mathcal{U})$
        \State $x_c, x_m \gets \text{RenderContentMask}(X_r,B_\mathrm{inside})$
        \State $x_r \gets \text{InpaintPatch}(X_r, x_c, x_m)$ 
           \Comment{Restore this patch}
        \State $\text{Paste}(x_r \text{ into } X_r \text{ at } (x_s,y_s,x_e,y_e))$
        \State \textbf{Mark} each box in $B_\mathrm{inside}$ \textbf{as restored} 
           $\bigl(\mathrm{restored\_flag}(b)\leftarrow\text{True}\bigr)$
    \EndFor
\EndWhile
\State \Return $X_r$
\end{algorithmic}
\end{algorithm*}

\end{document}